\documentclass[letterpaper]{article} 
\usepackage{aaai2026}  
\usepackage{times}  
\usepackage{helvet}  
\usepackage{courier}  
\usepackage[hyphens]{url}  
\usepackage{graphicx} 
\usepackage{natbib}  
\usepackage{caption} 
\usepackage{algorithm}
\usepackage{algorithmic}
\usepackage{makecell}
\usepackage{longtable}
\usepackage{newfloat}
\usepackage{listings}
\DeclareCaptionStyle{ruled}{labelfont=normalfont,labelsep=colon,strut=off} 
\floatstyle{ruled}
\newfloat{listing}{tb}{lst}{}
\floatname{listing}{Listing}
\usepackage{graphics} 
\usepackage{epsfig} 
\usepackage{mathptmx} 
\usepackage{mathrsfs}
\usepackage{times} 
\usepackage{amsmath} 
\usepackage{amssymb}  
\usepackage{multirow}    
\usepackage{booktabs}    
\usepackage{array}       
\usepackage{adjustbox}   
\usepackage{caption}     
\usepackage{cite}
\usepackage{mathrsfs}
\usepackage{pifont}
\usepackage[T1]{fontenc}

\DeclareMathAlphabet{\mathcal}{OMS}{cmsy}{m}{n}

\title{Zero-Shot Robotic Manipulation via 3D Gaussian Splatting-Enhanced Multimodal Retrieval-Augmented Generation}
\author{
    Zilong Xie\textsuperscript{\rm 1}\equalcontrib,
    Jingyu Gong\textsuperscript{\rm 1, \rm 2, \rm 3}\equalcontrib,
    Xin Tan\textsuperscript{\rm 1, \rm 2},
    Zhizhong Zhang\textsuperscript{\rm 1, \rm 3}\thanks{Corresponding authors},
    Yuan Xie\textsuperscript{\rm 1, \rm 2}
}
\affiliations{
    \textsuperscript{\rm 1}School of Computer Science and Technology, East China Normal University, Shanghai, China\\
    \textsuperscript{\rm 2}Chongqing Key Laboratory of Precision Optics, Chongqing Institute of East China Normal University, Chongqing, China\\
    \textsuperscript{\rm 3}Shanghai Key Laboratory of Computer Software Evaluating and Testing, Shanghai, China\\
    zilong6037@gmail.com, \{jygong, xtan, zzzhang, yxie\}@cs.ecnu.edu.cn
}

\begin{document}

\maketitle

\begin{abstract}

Existing end-to-end approaches of robotic manipulation often lack generalization to unseen objects or tasks due to limited data and poor interpretability. While recent Multimodal Large Language Models (MLLMs) demonstrate strong commonsense reasoning, they struggle with geometric and spatial understanding required for pose prediction. In this paper, we propose RobMRAG, a 3D Gaussian Splatting-Enhanced Multimodal Retrieval-Augmented Generation (MRAG) framework for zero-shot robotic manipulation. Specifically, we construct a multi-source manipulation knowledge base containing object contact frames, task completion frames, and pose parameters. During inference, a Hierarchical Multimodal Retrieval module first employs a three-priority hybrid retrieval strategy to find task-relevant object prototypes, then selects the geometrically closest reference example based on pixel-level similarity and Instance Matching Distance (IMD). We further introduce a 3D-Aware Pose Refinement module based on 3D Gaussian Splatting into the MRAG framework, which aligns the pose of the reference object to the target object in 3D space. The aligned results are reprojected onto the image plane and used as input to the MLLM to enhance the generation of the final pose parameters. Extensive experiments show that on a test set containing 30 categories of household objects, our method improves the success rate by 7.76\% compared to the best-performing zero-shot baseline under the same setting, and by 6.54\% compared to the state-of-the-art supervised baseline. Our results validate that RobMRAG effectively bridges the gap between high-level semantic reasoning and low-level geometric execution, enabling robotic systems that generalize to unseen objects while remaining inherently interpretable.
\end{abstract}

\begin{links}
    \link{Code}{https://github.com/XieZilongAI/RobMRAG}
\end{links}


\begin{figure*}[t]
    \centering
    \includegraphics[width=\textwidth]{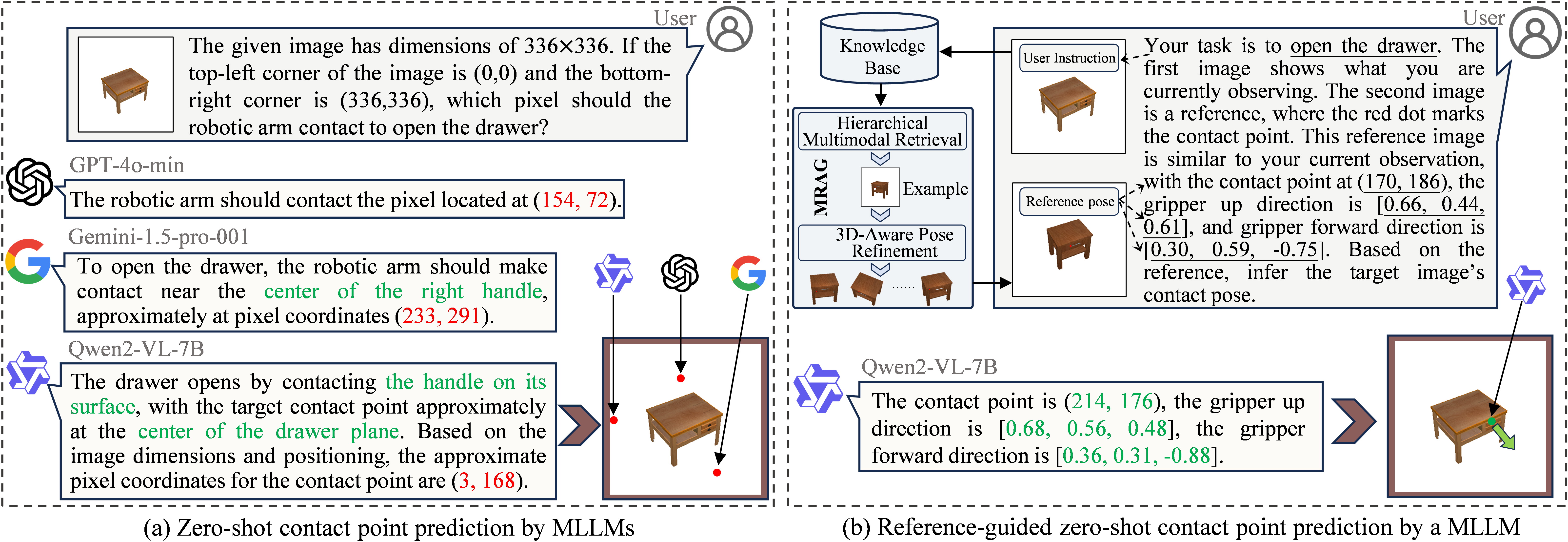}
    \caption{(a) Although multimodal large language models (MLLMs) possess the commonsense knowledge that opening a drawer requires grasping the handle and can roughly localize it, the predicted 2D contact points consistently fall outside the table itself, let alone on the handle. (b) We incorporate contact pose prediction using a multimodal retrieval-augmented generation (MRAG) framework. By retrieving relevant manipulation examples, our method enables more accurate inference of the contact pose on the target object.
    }
    \label{fig:introduction}
\end{figure*}

\section{Introduction}

Robotic manipulation fundamentally requires precise interaction with diverse objects, where accurate prediction of contact points and 3D poses is critical for task success. Without reliable estimation of object poses, robots are often forced into trial-and-error adjustments, resulting in inefficiency and failure~\cite{zhang2024pivot}. Therefore, robots must possess strong reasoning and generalization capabilities to accurately predict low-level action parameters and adapt to dynamic environments.

Although existing works~\cite{geng2023partmanip,geng2023gapartnet,geng2023rlafford} have explored end-to-end learning for grasp pose prediction, their generalization remains limited due to constrained data scale and the black-box nature of deep models, which lack commonsense reasoning capabilities. Recent advances in Multimodal Large Language Models (MLLMs)~\cite{zhang2023llama,li2022blip,li2024seed} have demonstrated promising performance in vision-language understanding and cross-modal reasoning. Several efforts~\cite{huang2023voxposer,brohan2023rt,li2024manipllm} utilize MLLMs for high-level instruction generation. For example, ManipLLM~\cite{li2024manipllm} employs chain-of-thought prompting and multi-task finetuning to guide category-level tasks. However, current MLLMs still lack sufficient understanding of geometric structure and spatial layout, limiting their effectiveness in precise grasp pose prediction, as shown in Figure~\ref{fig:introduction}(a).

Moreover, zero-shot generalization is a key capability for building general-purpose agents, enabling robots to manipulate unseen objects and adapt to varying embodiments and environments without task-specific training. Existing methods often rely on expert demonstrations~\cite{vuong2023open,khazatsky2024droid} or transfer from human interaction data such as HOI~\cite{liu2022hoi4d,grauman2022ego4d,luo2022learning} and internet videos~\cite{chen2025vidbot}. RAM~\cite{kuang2024ram} performs zero-shot manipulation via cross-domain affordance retrieval, yet its 2D-to-3D lifting pipeline is brittle under viewpoint or geometric shifts, and the MLLM's reasoning remains outside the geometric core, limiting pose precision in complex scenes.

To address these challenges, we propose RobMRAG, a 3D Gaussian Splatting-Enhanced Multimodal Retrieval-Augmented Generation (MRAG) framework for zero-shot robotic manipulation (see Figure~\ref{fig:introduction}(b)). First, a multi-source manipulation knowledge base is built from simulation, robotic dataset, and Internet data, containing rich multimodal information such as object contact frames, task-completion frames, and contact poses. During inference, a hierarchical multimodal retrieval module first uses a three-priority hybrid retrieval strategy to find task-relevant operation prototypes. Then, pixel-level cosine similarity and Instance Matching Distance (IMD) identify the closest geometric reference. Within the MRAG framework, a 3D Gaussian Splatting-based pose refinement module is introduced to perform rigid transformations on retrieved reference poses for precise 3D alignment with target objects. The aligned poses are then reprojected onto the 2D image plane and fed into the MLLM to guide accurate manipulation pose generation. Extensive experiments show that RobMRAG outperforms state-of-the-art methods under both zero-shot and supervised settings, effectively bridging task understanding and concrete operation for open-world robotic manipulation.

The key contributions of this work are summarized as follows:
\begin{itemize}
    \item We propose a Multimodal Retrieval-Augmented Generation (MRAG) framework for zero-shot robotic manipulation, which enables manipulation on unseen objects through a multi-source knowledge base.
    
    \item We integrate a 3D-Aware Pose Refinement module into the MRAG framework, enabling precise pose alignment between reference and target objects, thereby enhancing the geometric consistency of the retrieved results.
    
    \item Experimental results demonstrate that, on a test set comprising 30 categories of household objects, the proposed method achieves a 7.76\% improvement in success rate compared with the SOTA zero-shot baseline, and a 6.54\% improvement compared with the SOTA supervised baseline.
\end{itemize}

\section{Related Works}

\subsection{Zero-shot Robotic Manipulation}
Zero-shot robotic manipulation enables robots to perform novel tasks without task-specific training. Existing methods fall into two categories. The first attempts to directly learn transferable manipulation policies from human videos~\cite{bharadhwaj2024towards,chang2023look,bharadhwaj2023zero,grauman2022ego4d,xu2023xskill,chen2025vidbot}, either through imitation learning by extracting intermediate representations such as hand trajectories and poses, or by constructing reward functions for reinforcement learning. However, these methods often rely on manually collected video datasets~\cite{smith2019avid,xiong2021learning} or require online fine-tuning~\cite{bahl2022human}, which limits their applicability in real-world scenarios. The second category builds databases of diverse manipulation demonstrations~\cite{nguyen2022affordance,ju2024robo,kuang2024ram}, retrieving demonstrations similar to the current task and adapting them to robot execution. Yet, the effectiveness of such methods heavily depends on the coverage of the demonstration database, which limits their generalization to novel tasks or unseen scenes.

In contrast, our method explicitly bridges the inherent gap between task-level intent and instance-level geometric parameters by constructing a multimodal manipulation knowledge base and integrating Hierarchical Multimodal Retrieval with a 3D-Aware Pose Refinement module.

\subsection{Multimodal Large Language Models for Robotics}
Multimodal Large Language Models (MLLMs) unify visual perception, language understanding, and action planning into a single framework~\cite{zhang2023llama, li2022blip}. Models like RT-2~\cite{brohan2023rt} and RoboMamba~\cite{liu2024robomamba} leverage large-scale pretraining to translate natural language into executable control commands. Recent works~\cite{zhen20243d, kim2024openvla, yue2024deer} adopt Transformer-based architectures for grounding semantics to actions. To improve adaptability in dynamic environments, while VoxPoser~\cite{huang2023voxposer} introduces language-driven affordance reasoning, dynamically constructing 3D semantic value maps to guide manipulation. Beyond task-specific designs, foundation models such as PaLM-E~\cite{driess2023palm} and Flamingo~\cite{alayrac2022flamingo} further extend cross-modal reasoning capabilities. ManipLLM~\cite{li2024manipllm} further enhances generalization and stability in grasp pose prediction by incorporating adapters and chain-of-thought reasoning mechanisms. 

Building upon this foundation, we combine multimodal retrieval-augmented generation with MLLMs, significantly improving zero-shot generalization in robotic manipulation tasks.

\subsection{Multimodal Retrieval-Augmented Generation}
Retrieval-Augmented Generation (RAG) mitigates limitations of Large Language Models (LLMs) like outdated knowledge and hallucination by leveraging external memory~\cite{lewis2020retrieval, guu2020retrieval}. Extending to multimodal domains, Multimodal RAG (MRAG), which integrates both visual and textual inputs, has become a prominent research direction~\cite{chen2022murag, ma2024multi, bonomo2025visual, liu2025siqa, yu2025mramg, liu2025hm}. Representative MRAG approaches, such as MuRAG~\cite{chen2022murag}, M2RAG~\cite{ma2024multi}, and MRAMG~\cite{yu2025mramg}, typically enhance the quality of generation in tasks like visual question answering by retrieving image-text pairs from external memory, overcoming conventional RAG's limitations in visual understanding. Meanwhile, the incorporation of graph neural networks and knowledge graphs~\cite{dong2024advanced, edge2024local, guo2024lightrag, liu2025aligning, wu2024medical}, exemplified by GraphRAG~\cite{edge2024local} and LightRAG~\cite{guo2024lightrag}, captures complex cross-modal relations for better semantic reasoning. 

In this work, we integrate MRAG techniques with MLLMs in the domain of robotic manipulation, enabling high-precision prediction of operational poses.

\begin{figure*}[t]
    \centering
    \includegraphics[width=\textwidth]{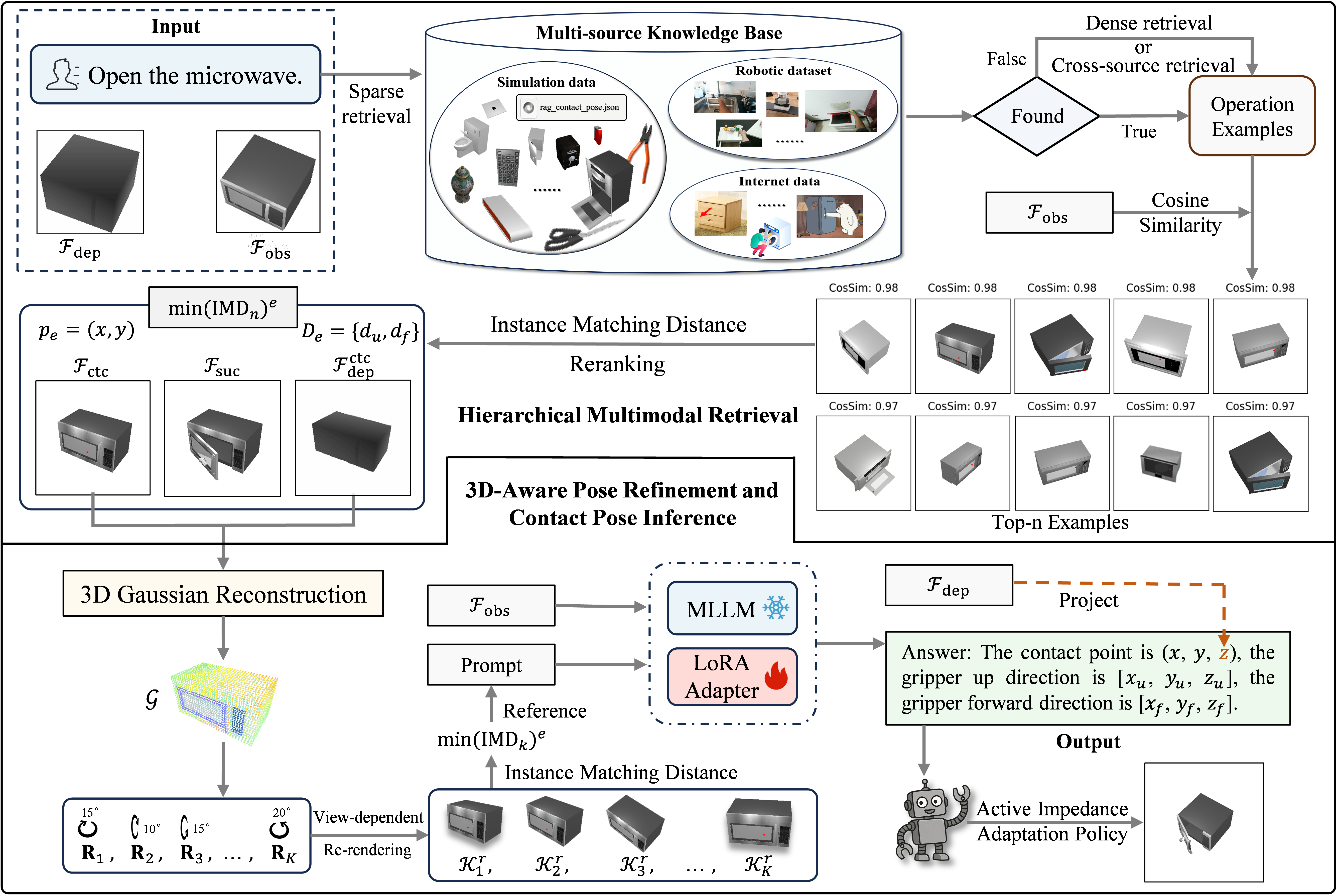}
    \caption{
     Overview of the RobMRAG: a 3D Gaussian splatting-enhanced multimodal retrieval-augmented generation framework.
    }
    \label{fig:framework}
\end{figure*}

\section{Method}

Our goal is to develop a framework for zero-shot robotic manipulation that can accurately and efficiently respond to user instructions in unseen environments. To this end, as illustrated in Figure~\ref{fig:framework}, we present RobMRAG, a 3D Gaussian Splatting-Enhanced Multimodal Retrieval-Augmented Generation framework for zero-shot robotic manipulation. In the following sections, we elaborate on the details of each component of RobMRAG.

\subsection{Multi-Source Knowledge Base Construction}
\label{sec:kb}

We construct a multimodal manipulation knowledge base that integrates data from simulation,  real-world or synthetic robotic dataset, and the Internet. Each manipulation case contains one or two key action frames: the object contact frame $\mathcal{F}_{\mathrm{ctc}}$, capturing the moment when the robot (or human) correctly makes contact with the object and initiates the manipulation, and the task-completion frame $\mathcal{F}_{\mathrm{suc}}$, showing the final successful state. It also includes a corresponding instruction text \(I_e\) describing the task. Specifically, for simulation data, we directly obtain precise 6D grasp parameters: the 2D contact point \(p_e=(x,y)\) and end-effector directions \(D_e=(d_u,d_f)\), where the gripper up direction \(d_u=(x_u,y_u,z_u)\). These data are collected from multiple successful executions across various object categories and manipulation types.

Real-world data are primarily sourced from human-object interaction video datasets (e.g.\ HOI4D~\cite{liu2022hoi4d}), where 2D contact points and motion directions are extracted via hand-keypoint detection and annotated as arrows on the contact-frame images. Internet data offer a more diverse range of manipulation scenarios, including examples extracted from instructional videos and animations. Although these cases often lack accurate 3D parameters, we apply a semi-automatic annotation pipeline to generate 2D grasp sketches. Ultimately, we construct a comprehensive multi-source manipulation knowledge base 
\(\mathcal{K} = \{(I_e,\;\mathcal{F}_{\mathrm{ctc}},\;\mathcal{F}_{\mathrm{suc}},\;p_e,\;D_e)\}\), 
which furnishes rich cross-domain priors for subsequent retrieval augmentation.

\subsection{Hierarchical Multimodal Retrieval}
\label{sec:retrieval}
Our framework employs a hierarchical retrieval strategy to identify the optimal reference example from the knowledge base. This process cascades from high-level textual semantics down to visual similarity and, finally, to fine-grained geometric matching, ensuring a precise instruction-to-example alignment.

\noindent
\textbf{Textual Semantic Retrieval.} The first layer operates on the textual modality, utilizing a three-priority hybrid strategy to find semantically relevant candidates. Given a natural language instruction \(I=\{w_i\}_{i=1}^L\), this strategy first executes its top priority: using sparse retrieval (BM25) to find manipulation examples with matching object names, primarily targeting our high-quality simulation data source. If no direct match is found (e.g., for a novel object category not in \(\mathcal{K}\)), the system defaults to its second priority: dense retrieval based on semantic embeddings. At this stage, the instruction text is projected into a dense vector by a pre-trained language encoder, and the cosine similarity \(S_{\text{text}}\) between semantic embeddings is then computed:
\begin{equation}
S_{\text{text}}(I, I_e)
= \frac{\hat{I}\cdot \hat{I}_e}{\|\hat{I}\|\,\|\hat{I}_e\|}.
\end{equation}
Here, \(\hat{I} = \mathrm{Enc}(I)\) and \(\hat{I}_e = \mathrm{Enc}(I_e)\) are the semantic embeddings of the instruction and an example text, respectively. As a final fallback, should the top similarity score fall below a threshold $\tau_{den}$, the third priority is triggered: the retrieval scope is expanded to our broader robotic dataset and Internet data sources. This textual retrieval stage yields a candidate set of examples that are conceptually aligned with the given task.

\noindent
\textbf{Visual Similarity Filtering.} Once the candidate set is obtained, the second layer refines it based on coarse visual similarity. The system computes the cosine similarity \(S_{\text{CLIP}}\) between the current observation image $\mathcal{F}_{\mathrm{obs}}$ and each candidate's contact frame $\mathcal{F}_{\mathrm{ctc}}$ using a CLIP image encoder, selecting the top-\(n\) visually similar samples:
\begin{equation}
S_{\text{CLIP}}(\mathcal{F}_{\mathrm{obs}}, \mathcal{F}_{\mathrm{ctc}}) = \frac{f_{\text{obs}} \cdot f_{\text{ctc}}}{\| f_{\text{obs}} \| \| f_{\text{ctc}} \|}
\end{equation}
where \( f_{\mathrm{obs}}\) and \( f_{\mathrm{ctc}}  \) are the visual feature
vectors extracted by CLIP from the \( \mathcal{F}_{\mathrm{obs}} \) and \( \mathcal{F}_{\mathrm{ctc}} \), respectively. This step effectively filters out candidates that, while semantically related, are visually distinct from the current scene.

\noindent
\textbf{Geometric Matching.} In the final layer, we perform fine-grained geometric matching on the remaining candidates to find the best reference prototype. We use Instance Matching Distance (IMD) to measure the geometric discrepancy between the observed object and each candidate by accounting for local feature consistency:
\begin{multline}
\text{IMD}(\mathcal{F}_{\mathrm{obs}}, \mathcal{F}_{\mathrm{ctc}}, \mathcal{M}_{\mathrm{obs}}) \\
= \sum_{p \in \mathcal{M}_{\mathrm{obs}}} \| \mathrm{F}^{\text{obs}}(p) - \text{NN}(\mathrm{F}^{\text{obs}}(p), \mathrm{F}^{\text{ctc}}) \|_2
\end{multline}
where $\mathcal{M}_{\mathrm{obs}}$ is the observed instance mask, $\mathbf{F}(p)$ is the dense feature vector at pixel $p$, and $\text{NN}(\cdot)$ finds the nearest neighbor match. The example with the minimum IMD is selected. If this score is below an empirically-determined threshold $\tau_{\text{IMD}}$, it is used directly as the final reference. Otherwise, it serves as a geometric prior for the subsequent pose refinement module.

\subsection{3D-Aware Pose Refinement }
\label{sec:pose}
For reference examples requiring further alignment, our framework employs a 3D-aware pose refinement module. We first generate a 3D Gaussian Splatting (3DGS)~\cite{kerbl20233d, gong2021omni, gboundary} representation from the reference's RGB-D image using a pre-trained generative model TRELLIS~\cite{xiang2025structured}. This allows us to re-render the object from the single input viewpoint. The core formulation expresses the object surface through a differentiable collection of Gaussian distributions $\mathcal{G} = \{(\mu_i, \Sigma_i, c_i)\}_{i=1}^N$ where $\mu_i \in \mathbb{R}^3$ denotes the Gaussian center position, $\Sigma_i \in \mathbb{R}^{3\times3}$ controls the spatial distribution, and $c_i$ represents the color attributes.

To eliminate the viewpoint discrepancy, our method applies a predefined set of small-angle rotational transformations $\{\mathbf{R_k}\}_{k=1}^{K}$ to the reference grasp pose. This strategy is sufficient as the preceding IMD filtering has already ensured a high degree of initial geometric alignment. We first project the 2D reference contact point $p_e$ to 3D space $p_e'$ using depth information, then transform both the 3D reference contact point and orientation:
\begin{equation}
\begin{aligned}
p_k' &= \mathbf{R_k} p_e' ,  &D_k^r = D_e \otimes \mathrm{Quat}(\mathbf{R_k})
\end{aligned}
\end{equation}
where $\otimes$ denotes quaternion multiplication. The differentiable rendering of 3DGS generates corresponding contact frames $\{\mathcal{F}_{\mathrm{ctc}}^r\}$ for each candidate pose after reprojecting the 3D contact points to the 2D image plane:
\begin{equation}
p_k^r = \pi(\mathbf{M}[\mathbf{R_k} | t]p_k')
\end{equation}
where $\pi(\cdot)$ is the camera projection function, $\mathbf{M}$ is the intrinsic matrix, and $t$ denotes the translational compensation. Finally, we obtain the re-rendered reference set \(\{\mathcal{K}_k^{r}\}_{k=1}^{K}\) and retain the pose whose frames yield the lowest $\mathrm{IMD}_k$ as the final reference inputs for the MLLM.

\subsection{Optional LoRA Fine-Tuning and Inference}
\label{sec:lora}
The framework employs a joint loss function combining Masked Language Modeling (MLM) with pose regression tasks, achieving precise grasp pose prediction through Low-Rank Adaptation (LoRA) of the multimodal large language model.

MLM randomly masks grasp parameters in input text, forcing the model to infer masked numerical characters from context. The loss function is defined as character-level cross-entropy at masked positions:
\begin{equation}
\mathcal{L}_{\text{MLM}} = -\sum_{i \in \mathcal{M}} \sum_{c \in \mathcal{C}_i} \log P(c \mid w_{\setminus \mathcal{M}}, \mathcal{F}_{\text{obs}}, \mathcal{F}_{\text{ctc}})
\end{equation}
where $\mathcal{M}$ denotes the set of masked positions, $\mathcal{C}_i$ represents the character sequence at the $i$-th masked position, $w_{\setminus M}$ indicates the unmasked text context, and $\mathcal{F}_{\text{obs}}$/$\mathcal{F}_{\text{ctc}}$ are the observation and reference images respectively.

The fine-tuning process supervises structured outputs with contact point $(x,y)$ constrained by mean squared error and end-effector directions $(d_u, d_f)$ regularized through cosine similarity:
\begin{equation}
\mathcal{L}_{\text{pose}} = \lambda_1 \|(x,y) - (\hat{x},\hat{y})\|_2^2 + \lambda_2 (2 - d_u \cdot \hat{d}_u - d_f \cdot \hat{d}_f)
\end{equation}
where $\lambda_1$, $\lambda_2$ are balance weights, $\hat{d}_u$ and $\hat{d}_f$ denote predicted direction vectors. The total loss combines both components: $\mathcal{L} = \mathcal{L}_{\text{MLM}} + \mathcal{L}_{\text{pose}}$.

During inference, the framework first retrieves the most task-relevant reference through Hierarchical Multimodal Retrieval and 3D-Aware Pose Refinement module, which are then provided as additional inputs to the MLLM for generating the operational pose. The final output adopts a structured format containing normalized 2D contact point and two 3D end-effector directions (representing gripper up and forward directions). 2D contact point is projected to 3D manipulation space via depth map  $\mathcal{F}_{\mathrm{dep}}$, constructing complete end-effector poses. We apply active impedance adaptation policy~\cite{li2024manipllm} to adjust movement direction.

\section{Experiments}

\subsection{Datasets and Evaluation Metrics}
We construct interactive manipulation environments based on the SAPIEN simulator and the PartNetMobility dataset~\cite{xiang2020sapien}, which provides part-level motion annotations for a wide range of object categories. Following the baseline setup~\cite{li2024manipllm}, simulated manipulation is performed using a Franka Panda robotic arm equipped with a suction gripper, rendered through the VulkanRenderer engine. To generate training data, we randomly sample contact points on movable parts and use the inverse surface normal as the initial operation direction, focusing on pull-type actions to ensure consistency between the motion and manipulation directions. In total, we collect approximately 20,000 successful manipulation samples across 20 object categories as our training set. Following the same procedure, we generate the test set, which is further divided into a Test Seen Split, containing objects of seen categories but with different instances, and a Test Unseen Split, containing entirely novel object categories.

The primary evaluation metric is manipulation average success rate (ASR)~\cite{li2024manipllm}, defined as the proportion of successful trials exceeding a displacement threshold $\delta$. We use two levels: $\delta=0.01$ for verifying initial pose prediction, and $\delta=0.1$ for assessing sustained motion. Active impedance control is applied to adapt motion direction in response to interaction uncertainties.

\subsection{Competitor Methods}
To validate our Multimodal Retrieval-Augmented Generation (MRAG) framework, we compare it against six robotic manipulation methods under identical train/test splits and end-effector configurations (suction grippers substituted for original parallel grippers in baselines). We compare our method with the following baselines: Where2Act~\cite{mo2021where2act} predicts pixel-level affordances for generalization; UMPNet~\cite{xu2022universal} generates 6DoF actions from monocular input via temporal causality; Flowbot3D~\cite{eisner2022flowbot3d} infers 3D motion fields from point clouds for planning; Implicit3D~\cite{zhong20233d} extends Transporter to 3D using spatio-temporal keypoints; RAM~\cite{kuang2024ram} is a zero-shot framework lifting retrieved 2D affordances to 3D actions; ManipLLM~\cite{li2024manipllm} uses MLLM chain-of-thought for planning and pose prediction; CrayonRobo~\cite{li2025object} is a prompt-driven VLA model using 2D visual prompts on images to predict SE(3) actions.

For evaluation, we consider the following settings: \textit{Zero-shot}: Directly applies the method without task-specific training or fine-tuning. \textit{All}: After fine-tuning, the model is evaluated on the complete test set to assess its generalization capability in real-world scenarios. \textit{Local}: After fine-tuning, evaluation is performed only on samples where the contact point is correctly predicted, representing the theoretical upper bound given accurate contact point localization.

\subsection{Comparative Analysis}

\begin{table*}[t] 
\centering
\renewcommand{\arraystretch}{0.9}
\setlength{\extrarowheight}{0pt}

\normalsize  

\resizebox{\textwidth}{!}{

  \begin{tabular}{@{}l*{16}{c}@{}}
    \toprule
  \multirow{2}{*}{Method} & \multicolumn{16}{c}{Test Seen Split} \\
  \cmidrule(lr){2-17}
& \includegraphics[height=0.6cm, keepaspectratio]{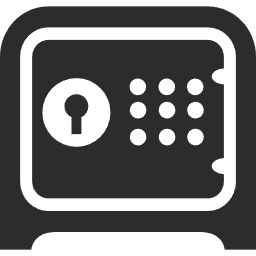} 
& \includegraphics[height=0.6cm, keepaspectratio]{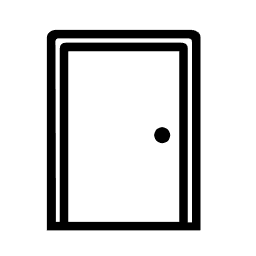} 
& \includegraphics[height=0.6cm, keepaspectratio]{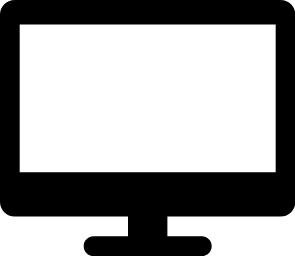} 
& \includegraphics[height=0.6cm, keepaspectratio]{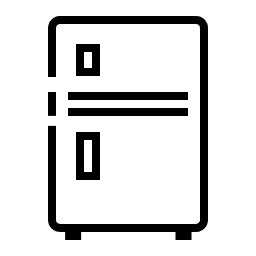}
& \includegraphics[height=0.6cm, keepaspectratio]{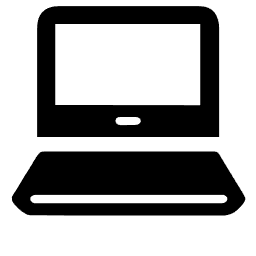} 
& \includegraphics[height=0.6cm, keepaspectratio]{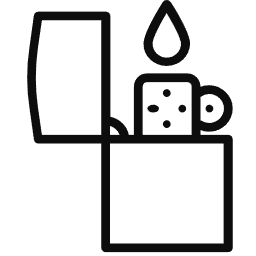} 
& \includegraphics[height=0.6cm, keepaspectratio]{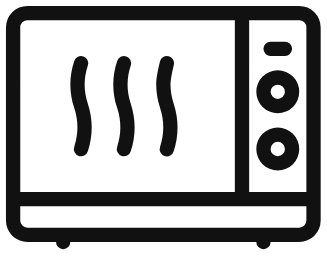} 
& \includegraphics[height=0.6cm, keepaspectratio]{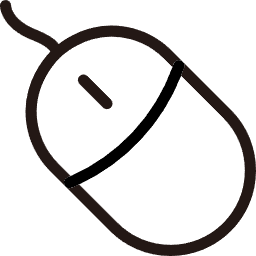} 
& \includegraphics[height=0.6cm, keepaspectratio]{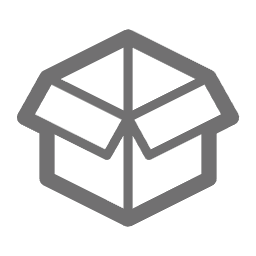} 
& \includegraphics[height=0.6cm, keepaspectratio]{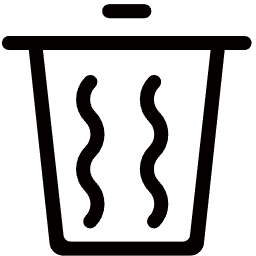} 
& \includegraphics[height=0.6cm, keepaspectratio]{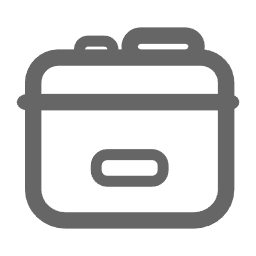} 
& \includegraphics[height=0.6cm, keepaspectratio]{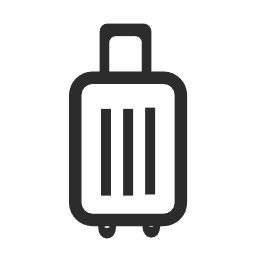} 
& \includegraphics[height=0.6cm, keepaspectratio]{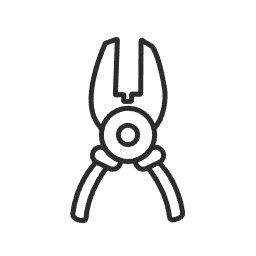} 
& \includegraphics[height=0.6cm, keepaspectratio]{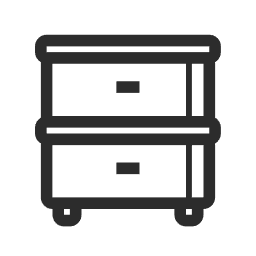} 
& \includegraphics[height=0.6cm, keepaspectratio]{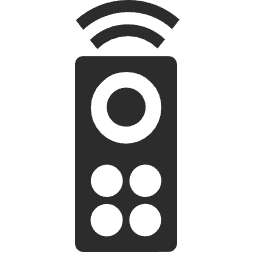} 
& \includegraphics[height=0.6cm, keepaspectratio]{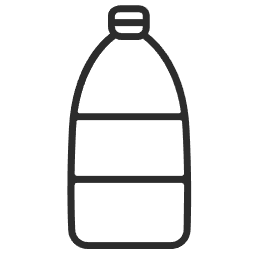} \\
\midrule
RAM~\cite{kuang2024ram} (\textit{Zero-shot})\textsuperscript{\dag} & 43.14 & 25.67 & 14.00 & 35.29 & 52.94 & 35.49 & 49.02 & 45.39 & 52.00 & 49.23 & 45.76 & 36.02 & 24.00 & 61.58 & 24.00 & 72.00 \\
CrayonRobo~\cite{li2025object} (\textit{Zero-shot})\textsuperscript{\dag} & 17.65 & 21.59 & 28.00 & 47.06 & 62.18 & 33.33 & 57.45 & 58.54 & 48.00 & 49.19 & 37.24 & 57.39 & 26.00 & 73.33 & 12.00 & 68.00 \\
Ours (\textit{Zero-shot}) & 62.75 & 47.45 & 20.00 & 66.67 & 50.98 & 17.65 & 59.18 & 72.05 & 64.00 & 43.14 & 49.02 & 38.00 & 22.00 & 67.85 & 18.00 & 78.00 \\
\midrule
Where2Act~\cite{mo2021where2act} & 26.00 & 36.00 & 19.00 & 27.00 & 23.00 & 11.00 & 15.00 & 47.00 & 14.00 & 24.00 & 13.00 & 12.00 & 56.00 & 68.00 & 7.00 & 40.00 \\
UMPNet~\cite{xu2022universal} & 46.00 & 43.00 & 15.00 & 28.00 & 54.00 & 32.00 & 28.00 & 56.00 & 44.00 & 40.00 & 10.00 & 23.00 & 18.00 & 54.00 & 20.00 & 42.00 \\
Flowbot3D~\cite{eisner2022flowbot3d} & 67.00 & 55.00 & 20.00 & 32.00 & 27.00 & 31.00 & 61.00 & 68.00 & 15.00 & 28.00 & 36.00 & 18.00 & 21.00 & 70.00 & 18.00 & 26.00 \\
Implicit3D~\cite{zhong20233d} & 53.00 & 58.00 & 35.00 & 55.00 & 28.00 & \textbf{66.00} & 58.00 & 51.00 & 52.00 & 57.00 & 45.00 & 34.00 & 41.00 & 54.00 & 39.00 & 43.00 \\
ManipLLM~\cite{li2024manipllm} & \textbf{68.00} & \textbf{64.00} & \textbf{36.00} & 77.00 & 43.00 & 62.00 & 65.00 & 61.00 & 65.00 & 52.00 & 53.00 & 40.00 & \textbf{64.00} & 71.00 & \textbf{60.00} & 64.00 \\

ManipLLM \textit{(All})\textsuperscript{\dag} & 60.78 & 51.57 & 46.00 & 52.94 & 37.25 & 33.33 & 63.51 & 62.56 & 46.00 & 43.14 & 25.49 & 54.00 & 42.00 & 65.83 & 22.00 & 78.00 \\
Ours (\textit{All}) & 66.67 & 51.56 & 22.00 & \textbf{82.27} & \textbf{96.08} & 18.73 & \textbf{82.27} & \textbf{87.12} & \textbf{66.00} & \textbf{65.85} & \textbf{75.62} & \textbf{56.32} & 32.00 & \textbf{77.21} & 22.00 & \textbf{96.00} \\
\midrule
ManipLLM (\textit{Local})\textsuperscript{\dag} & 72.73 & 67.89 & 47.83 & 63.41 & 44.96 & 64.00 & 75.37 & 82.43 & 63.16 & 54.29 & 26.00 & 84.62 & 63.64 & 72.07 & 29.04 & 80.80  \\
Ours (\textit{Local}) & 74.07 & 66.01 & 24.29 & 85.05 & 98.00 & 21.86 & 100.0 & 100.0 & 87.24 & 72.12 & 88.24 & 100.0 & 71.43 & 82.12 & 31.25 & 97.96  \\
\midrule
\multirow{2}{*}{Method} & \multicolumn{5}{c}{Test Seen Split} & \multicolumn{11}{c}{Test Unseen Split} \\
\cmidrule(lr){2-6} \cmidrule(lr){7-17}
& \includegraphics[height=0.6cm, keepaspectratio]{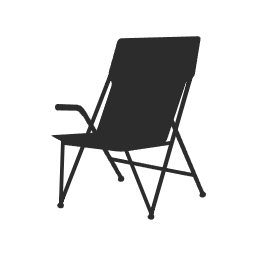} 
& \includegraphics[height=0.6cm, keepaspectratio]{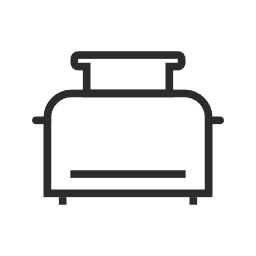} 
& \includegraphics[height=0.6cm, keepaspectratio]{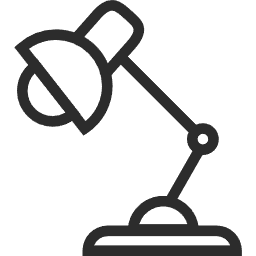} 
& \includegraphics[height=0.6cm, keepaspectratio]{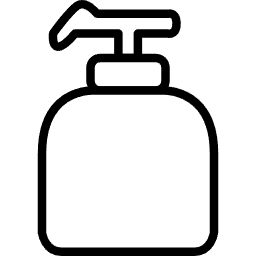} 
& \raisebox{0.6\height}{\textbf{AVG}}
& \includegraphics[height=0.6cm, keepaspectratio]{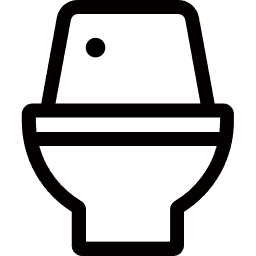} 
& \includegraphics[height=0.6cm, keepaspectratio]{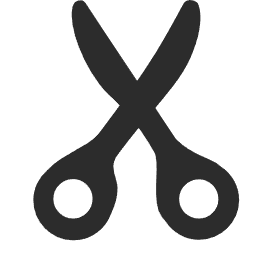} 
& \includegraphics[height=0.6cm, keepaspectratio]{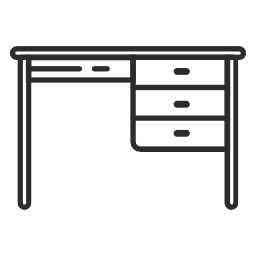} 
& \includegraphics[height=0.6cm, keepaspectratio]{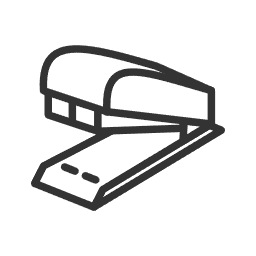} 
& \includegraphics[height=0.5cm, keepaspectratio]{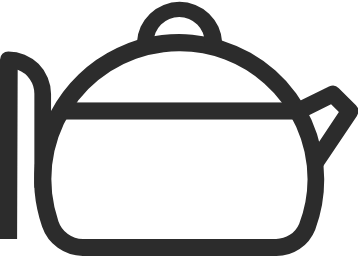} 
& \includegraphics[height=0.6cm, keepaspectratio]{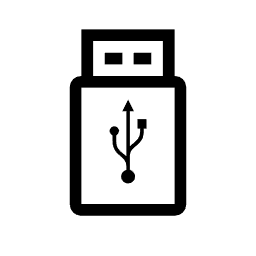} 
& \includegraphics[height=0.6cm, keepaspectratio]{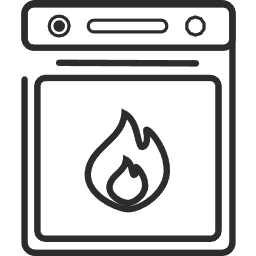} 
& \includegraphics[height=0.6cm, keepaspectratio]{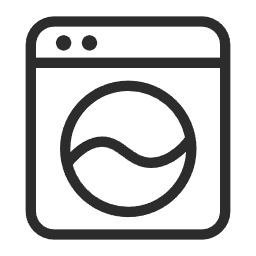} 
& \includegraphics[height=0.6cm, keepaspectratio]{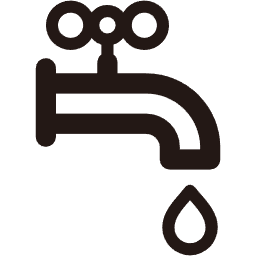} 
& \includegraphics[height=0.6cm, keepaspectratio]{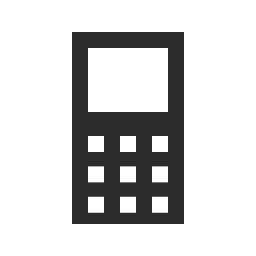} 
& \raisebox{0.6\height}{\textbf{AVG}}  \\
  \cmidrule(lr){1-6} \cmidrule(lr){7-17}
RAM~\cite{kuang2024ram} (\textit{Zero-shot})\textsuperscript{\dag} & 32.45 & 52.64 & 21.61 & 38.93 & 40.56 & 42.38 & 25.60 & 42.83 & 56.07 & 30.47 & 38.00 & 32.00 & 43.06 & 18.37 & 28.96 & 35.77 \\
CrayonRobo~\cite{li2025object} (\textit{Zero-shot})\textsuperscript{\dag} & 47.06 & 54.90 & 16.33 & 51.02 & 43.31 & 48.67 & 29.92 & 31.67 & 58.36 & 18.68 & 24.00 & 38.00 & 41.18 & 15.69 & 27.45 & 33.36 \\
Ours (\textit{Zero-shot}) & 48.98 & 76.14 & 15.69 & 67.47 & 49.25 & 60.78 & 25.49 & \textbf{82.75} & 68.38 & 28.69 & 18.00 & 40.00 & 64.71 & 13.73 & 32.76 & 43.53 \\
\midrule
Where2Act~\cite{mo2021where2act} & 13.00 & 18.00 & 13.00 & 40.00 & 26.00 & 18.00 & \textbf{35.00} & 38.00 & 28.00 & 5.00 & 21.00 & 17.00 & 20.00 & 15.00 & 15.00 & 21.00 \\
UMPNet~\cite{xu2022universal} & 22.00 & 33.00 & 26.00 & 64.00 & 35.00 & 42.00 & 20.00 & 35.00 & 42.00 & 29.00 & 20.00 & 26.00 & 28.00 & 25.00 & 15.00 & 28.00 \\
Flowbot3D~\cite{eisner2022flowbot3d} & 17.00 & 53.00 & 29.00 & 42.00 & 37.00 & 23.00 & 10.00 & 60.00 & 39.00 & 27.00 & 42.00 & 28.00 & 51.00 & 13.00 & 23.00 & 32.00 \\
Implicit3D~\cite{zhong20233d} & 27.00 & 65.00 & 20.00 & 33.00 & 46.00 & 45.00 & 17.00 & 80.00 & 53.00 & 15.00 & 69.00 & 41.00 & 31.00 & \textbf{30.00} & 31.00 & 41.00 \\
ManipLLM~\cite{li2024manipllm} & 41.00 & 75.00 & \textbf{44.00} & 67.00 & 56.00 & 38.00 & 22.00 & 81.00 & \textbf{86.00} & 38.00 & \textbf{85.00} & 42.00 & \textbf{83.00} & 26.00 & 38.00 & 51.00 \\

ManipLLM (\textit{All})\textsuperscript{\dag} & 24.49 & 64.78 & 31.37 & 58.92 & 48.20 & 29.41 & 17.65 & 80.27 & 78.51 & 29.03 & 72.00 & 28.00 & 66.67 & 15.69 & 36.20 & 45.34 \\
Ours (\textit{All}) & \textbf{72.92} & \textbf{83.51} & 20.84 & \textbf{68.06} & \textbf{62.15} & \textbf{65.21} & 17.84 & 79.45 & 78.21 & \textbf{56.09} & 42.00 & \textbf{86.00} & 67.84 & 25.65 & \textbf{57.13} & \textbf{57.54} \\
\midrule
ManipLLM (\textit{Local})\textsuperscript{\dag} & 35.90 & 73.36 & 53.85 & 69.74 & 61.25 & 41.67 & 19.09 & 87.50 & 85.28 & 37.34 & 85.29 & 43.33 & 76.67 & 20.77 & 40.35 & 53.73 \\
Ours (\textit{Local}) & 97.22 & 92.30 & 48.78 & 71.14 & 75.45 & 72.22 & 20.31 & 83.33 & 86.02 & 65.68 & 78.78 & 95.54 & 78.15 & 28.57 & 66.41 & 67.50 \\
   \bottomrule
  \end{tabular}%
  }
\caption{Comparisons of our method against baseline methods, \textsuperscript{\dag}indicates reproduced results (ASR, \%).}
\label{tab:results}
\end{table*}

As shown in Table~\ref{tab:results}, our framework's efficacy is first demonstrated in the \textit{zero-shot} setting, where RobMRAG achieves an average success rate of 43.53\% on the unseen split, surpassing the RAM baseline by 7.76 percentage points. This highlights the strength of our framework, which combines the advantages of MLLMs and MRAG. This design enables effective utilization of external knowledge bases, allowing the model to generalize to novel objects without task-specific training, as seen in high-performing tasks like \texttt{Table} (82.75\%). In the fine-tuned \textit{All} setting, our model's advantage becomes more pronounced: it achieves a 57.54\% average SR on unseen categories, which is 6.54 percentage points higher than the strongest baseline, ManipLLM. On certain tasks, our model reaches exceptionally high success rates, such as \texttt{Oven} (86.00\% vs. ManipLLM's 42.00\%) and \texttt{Laptop} (96.08\% vs. 43.00\%). In the \textit{Local} evaluation setting, where a perfect contact point is provided, our pose generation process still demonstrates superior robustness. Specifically, we observe average success rate improvements of 14.20\% on the seen split and 13.77\% on the unseen split. These gains are attributed to our 3D-Aware Pose Refinement module, which aligns retrieved examples in 3D space using Gaussian Splatting and reprojects them to provide the MLLM with more accurate geometric context. This enriched spatial information leads to more precise 6-DoF pose estimation.

Despite the strong overall performance, our analysis also identifies challenges. The first stems from the inherent properties of the evaluation dataset itself. For instance, some tasks exhibit low success rates due to severely degraded visual input; certain \texttt{Door} images are reduced to a few feature lines with nearly 80\% pixel loss, making robust feature extraction nearly impossible. Similarly, tasks like \texttt{Faucet} (25.65\% SR) suffer from ambiguous instructions in the dataset, where the prompt does not distinguish between "lever adjustment" and "spout repositioning." These dataset-level issues present a fundamental challenge to any vision-language-based method. The second challenge, however, reflects a genuine limitation of our current framework: manipulation of objects with minuscule interaction regions. The low SR on tasks requiring sub-centimeter precision, such as \texttt{Scissors} (17.84\%) and \texttt{Pliers} (32.00\%), indicates that our model still struggles with the fine-grained physical reasoning and exacting grasping point prediction necessary for such millimeter-scale operations.

\subsection{Ablation Studies}

\begin{table}[t]
\centering

\begin{tabular}{lcc}
\toprule
\textbf{Ablation Method} & \makecell{\textbf{Test} \\ \textbf{seen}} & \makecell{\textbf{Test} \\ \textbf{unseen}} \\
\midrule
\textit{Main Component Ablation} &  & \\
w/o Retrieval (only vision input) & 19.46 & 19.01 \\
w/ Textual Retrieval + CosSim only & 37.05 & 34.62 \\
w/ Textual Retrieval + IMD only & 45.55 & 41.27 \\
w/ Textual Retrieval + CosSim + IMD & 47.34 & 42.08 \\
\textbf{Full MRAG (w/ 3D Pose Align)} & \textbf{49.25} & \textbf{43.53} \\
\midrule
\textit{Different MLLM Backbones} &  & \\
Qwen2-VL-7B-Instruct (base) & 62.15 & 57.54 \\
LLaMA3.2-11B-Vision-Instruct & 62.93 & 58.18 \\
Qwen2.5-VL-7B-Instruct & \textbf{65.52} & \textbf{60.63} \\
\bottomrule
\end{tabular}
\caption{Ablation studies on different components of our MRAG framework (ASR, \%).}
\label{tab:ablation}
\end{table}

We conduct a series of ablation studies to evaluate the contributions of different components in the MRAG framework. As shown in Table~\ref{tab:ablation}, the first block presents the zero-shot performance of the MLLM without utilizing our proposed framework. We observe that using only visual input without incorporating any additional signals leads to poor performance (19.46\% seen / 19.01\% unseen). Introducing multimodal retrieval based on either cosine similarity or Instance Matching Distance (IMD) significantly improves the results, and combining both further enhances generalization capability. Building on this, we further incorporate the 3D-Aware Pose Refinement module, which yields the best zero-shot results (49.25\% seen / 43.53\% unseen), highlighting the crucial role of spatial grounding in improving retrieval performance.

We then evaluate the framework's performance with different MLLM backbones in the fine-tuned \textit{All} setting. As shown in the table, leveraging more advanced models consistently enhances results, demonstrating our framework's scalability. Specifically, upgrading the backbone from the base Qwen2-VL-7B-Instruct to Qwen2.5-VL-7B-Instruct achieves the overall peak performance of 65.52\% on the seen split and 60.63\% on the unseen split. Per-category success rates are detailed in the Appendix. These results clearly indicate that our RobMRAG architecture effectively capitalizes on the enhanced reasoning and visual understanding capabilities of more powerful MLLMs, establishing a robust and scalable foundation for robotic manipulation.

\begin{figure}[t]
    \centering
    \includegraphics[width=1.0\linewidth]{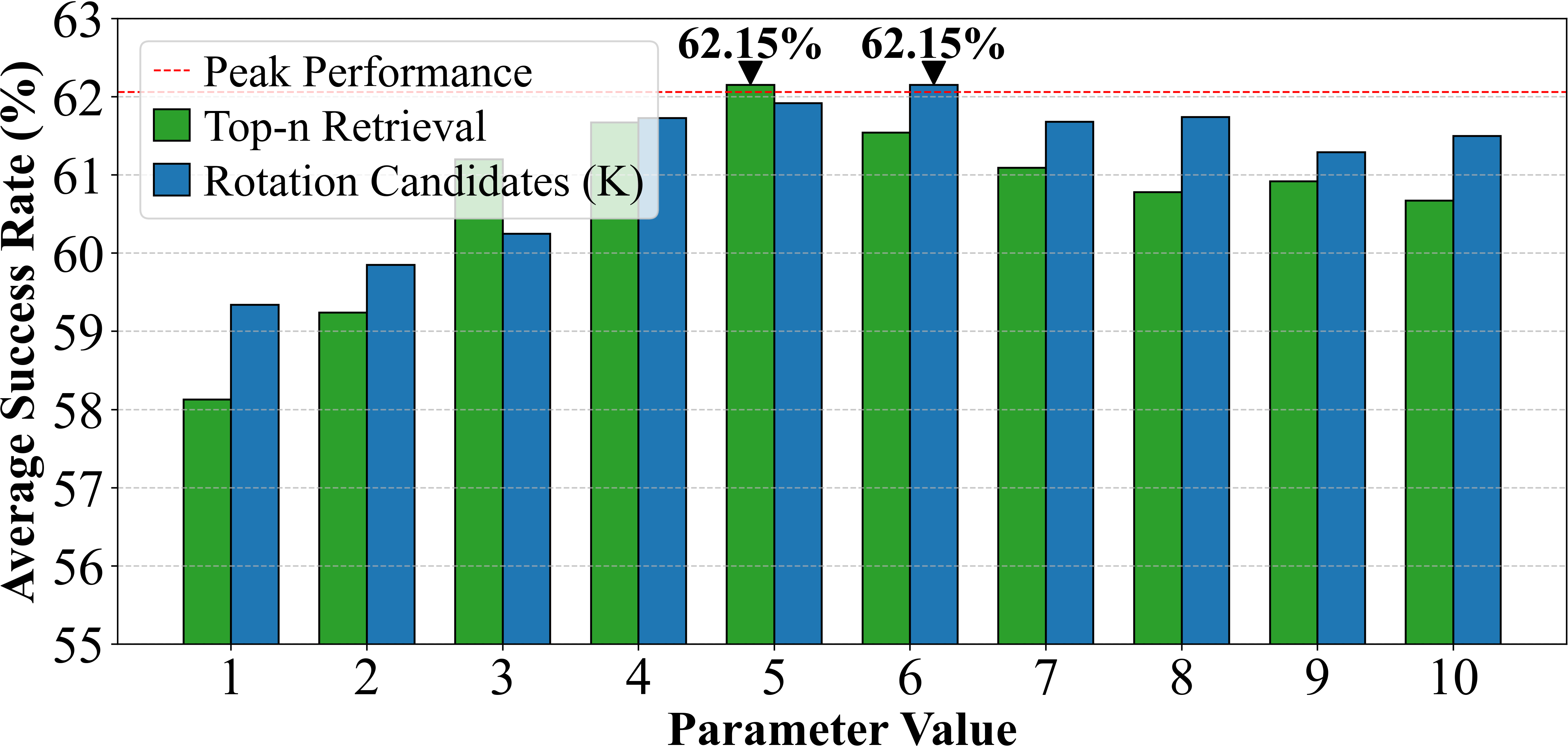}
    \caption{Ablation study on the number of retrieval candidates (Top-n) and rotation samples (\textit{K}).}
    \label{fig:ablation}
\end{figure}

We conducted controlled ablation studies on the retrieval candidate number (top-n) and rotation sampling count (\textit{K}) separately on the test seen split, as shown in Figure~\ref{fig:ablation}. Experiments reveal: peak success rate of 62.15\% occurs at top-n=5, with larger values degrading performance by 0.61--1.48\% due to geometric noise; optimal \textit{K}=6 achieves the best performance, validating small-angle rotation compensation. Both parameters exhibit an initial increase followed by a decline, demonstrating the necessity to balance search and sampling breadth with geometric consistency. Excessive values reduce success rates and increase computational overhead at the same time.

\subsection{Qualitative Results}

\begin{figure}[t]
    \centering
    \includegraphics[width=1.0\linewidth]{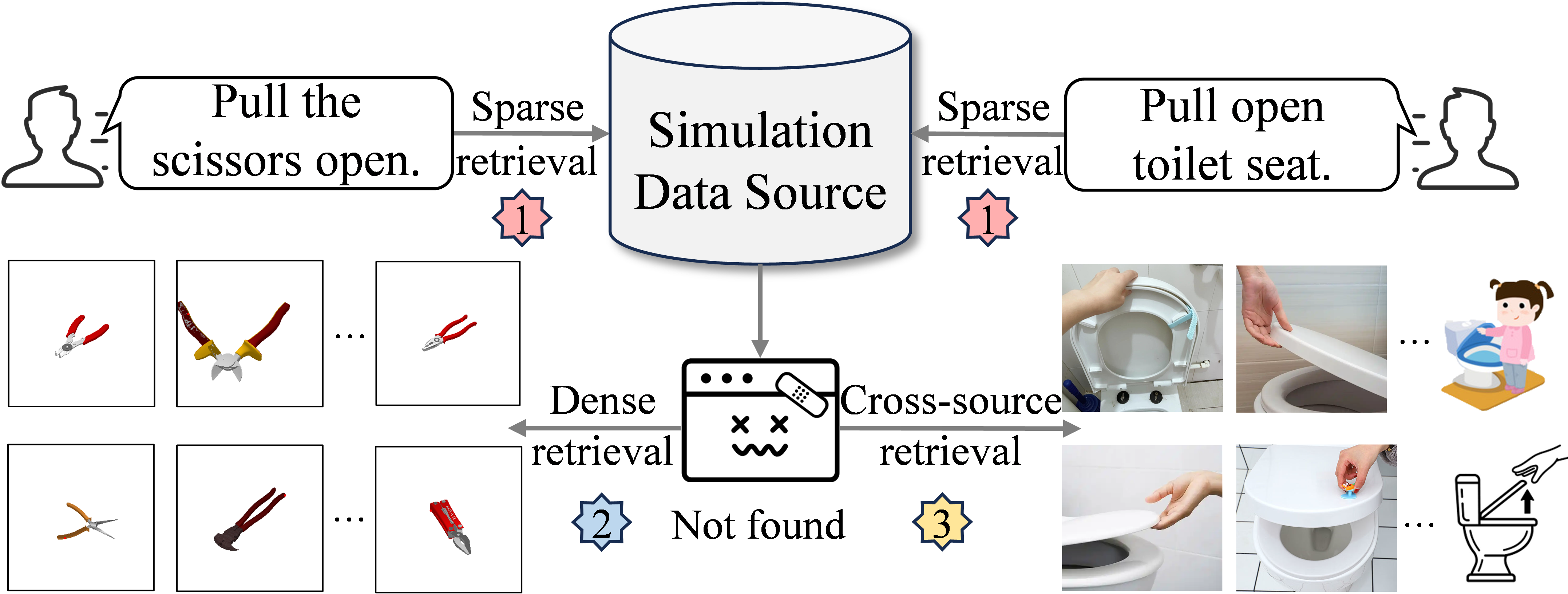}
    \caption{Visualization of the three-priority hybrid retrieval strategy: \ding{172} sparse retrieval (simulation), \ding{173} dense retrieval (simulation), and \ding{174} cross-source retrieval (robotic/Internet).}
    \label{fig:hybrid_retrieval}
\end{figure}

Figure~\ref{fig:hybrid_retrieval} presents qualitative examples demonstrating the robustness of our three-priority hybrid retrieval strategy. The left example showcases the second-priority fallback: for the target object \texttt{Scissors}, which is absent from our simulation dataset, the system uses dense retrieval to successfully match \texttt{Pliers} as a functionally analogous counterpart for affordance reference. The right example illustrates the third-priority case: given a \texttt{Toilet} with no direct or semantically close match in simulation, the strategy expands the search to robotic and Internet datasets to retrieve a valid reference.

\begin{figure}[t]
    \centering
    \includegraphics[width=1.0\linewidth]{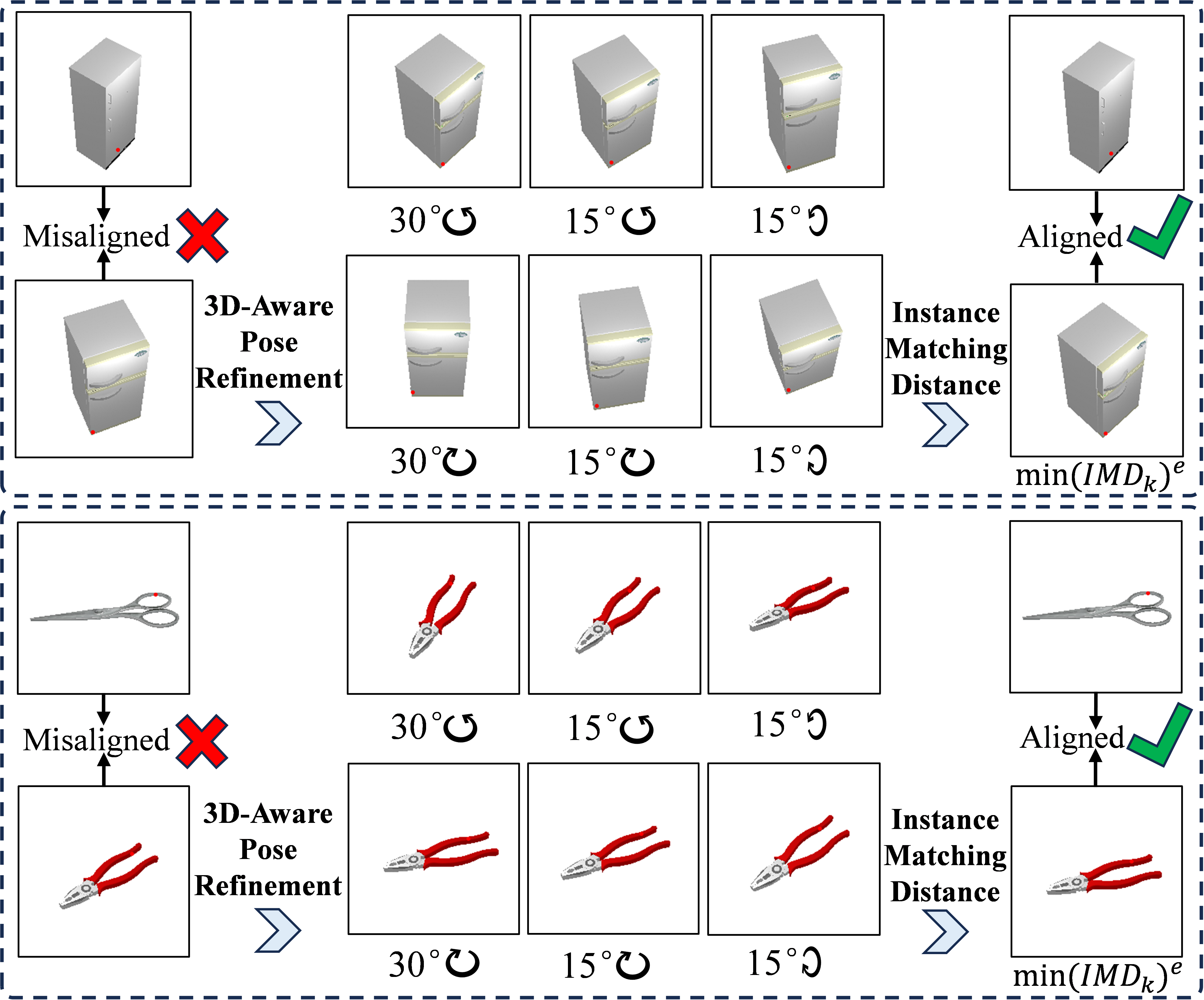}
    \caption{Visualization of 3D-Aware Pose Refinement.}
    \label{fig:3D_Aware}
\end{figure}

Figure~\ref{fig:3D_Aware} illustrates the geometric alignment process for reference poses using 3D-Aware Pose Refinement module. Due to viewpoint discrepancies between the retrieved reference pose and the observed target object, we employ 3D Gaussian Splatting to generate candidate views by applying multiple rotational transformations. The IMD is then recalculated for each candidate, and the pose with the minimal IMD is selected to achieve precise alignment.

\section{Conclusion}
We propose a novel robot manipulation framework based on Multimodal Retrieval-Augmented Generation (MRAG). By integrating a multi-source manipulation knowledge base, hierarchical multimodal retrieval, and a 3D-aware pose refinement module, our approach improves the average manipulation success rates by 7.76\% over state-of-the-art baselines in a zero-shot setting across various objects. Furthermore, with LoRA fine-tuning, the performance improves by up to 6.54\% compared to the baseline. While our method demonstrates strong capabilities in semantic understanding and pose adaptation, challenges remain in achieving millimeter-level precision and complex spatial reasoning. Future work will focus on enhancing fine-grained physical reasoning and spatial understanding.



\section*{Acknowledgements}
This work is supported by the National Natural Science Foundation of China (Grant No. 62176092, 62222602, 62302167, U23A20343, 62476090, 62502159), Natural Science Foundation of Shanghai (Grant No. 25ZR1402135), Shanghai Sailing Program (Grant No. 23YF1410500), Young Elite Scientists Sponsorship Program by CAST (Grant No. YESS20240780), the Chenguang Program of Shanghai Education Development Foundation and Shanghai Municipal Education Commission (Grant No. 23CGA34), Natural Science Foundation of Chongqing (Grant No. CSTB2023NSCQ-JQX0007,CSTB2023NSCQ-MSX0137, CSTB2025NSCQ-GPX0445), Open Project Program of the State Key Laboratory of CAD\&CG (Grant No. A2501), Zhejiang University, Open Research Fund of Key Laboratory of Advanced Theory and Application in Statistics and Data Science-MOE, ECNU.




\bigskip

\bibliography{aaai2026}

\end{document}